\documentclass[conference]{IEEEtran}

\usepackage{times}
\usepackage{epsfig}
\usepackage{graphicx}
\usepackage{amsmath}
\usepackage{amssymb}
\usepackage{booktabs}

\usepackage{adjustbox}
\usepackage{array}

\newcolumntype{R}[2]{%
  >{\adjustbox{valign=m, angle=#1,lap=\width-(#2)}\bgroup}%
  l%
  <{\egroup}%
}
\newcommand*\rot{\multicolumn{1}{R{90}{1em}}}

\newcommand{\etal}{\mbox{\emph{et al.\ }}}

\graphicspath{{./Figures/}}
\usepackage{array}
\usepackage[font=small,labelfont=bf,up,textfont=up]{caption}
\usepackage{subcaption}
\usepackage{multirow}

\newcommand{\ig}[2]{\includegraphics[width=#1]{#2} } 
\newlength{\sm}
\usepackage[pagebackref=true,breaklinks=true,letterpaper=true,colorlinks,bookmarks=false]{hyperref}

\begin{document}




\title{A Study and Comparison of Human and Deep Learning Recognition Performance \\ Under Visual Distortions}

\author{Samuel Dodge and Lina Karam\\
Arizona State University\\
{\tt\small \{sfdodge,karam\}@asu.edu}
}

\maketitle

\begin{abstract}
Deep neural networks (DNNs) achieve excellent performance on standard classification tasks. However, under image quality distortions such as blur and noise, classification accuracy becomes poor. In this work, we compare the performance of DNNs with human subjects on distorted images. We show that, although DNNs perform better than or on par with humans on good quality images, DNN performance is still much lower than human performance on distorted images. We additionally find that there is little correlation in errors between DNNs and human subjects. This could be an indication that the internal representation of images are different between DNNs and the human visual system. These comparisons with human performance could be used to guide future development of more robust DNNs.
\end{abstract}

\section{Introduction}
Recently deep neural networks (DNNs) have attained impressive performance in many fields such as image classification \cite{resnet}, semantic segmentation \cite{fcn}, and image compression \cite{google-comp}. The performance of these deep networks has begun to exceed human performance in many tasks. Human top-5 classification error rate on the large scale ImageNet dataset has been reported to be 5.1\% \cite{imagenet}, whereas a state-of-the-art neural network \cite{resnet} achieves a top-5 error rate of 3.57\%.

Most existing works assume that the input images are of good quality. However in many practical scenarios, images may be distorted. Image distortions can arise during acquisition or transmission. In acquisition, the image sensor can exhibit noise in low light conditions. Motion blur can occur if the camera is moving. In transmission, packet-loss could cause missing regions of the image, or missing frequencies, depending on how the image is encoded. As machine learning has begun to see popularity as a cloud-based service, these issues have become more relevant.

It has been shown that neural network performance decreases under image quality distortions \cite{dodge}. The degradation is particularly evident for additive noise or blur distortions. Given that networks equal or exceed human performance on undistorted images, it is interesting to ask: do networks still achieve equal or greater performance as compared to humans on distorted images? If human and DNN performance are similar on distorted images, then distorted images may be inherently difficult to recognize. Conversely, if it is shown that humans exceed DNN performance on distorted images, then there is some representational capacity present in the human visual system that is lacking in DNNs.

\begin{figure}[!tb]
  \centering
  \includegraphics[width=0.5\textwidth]{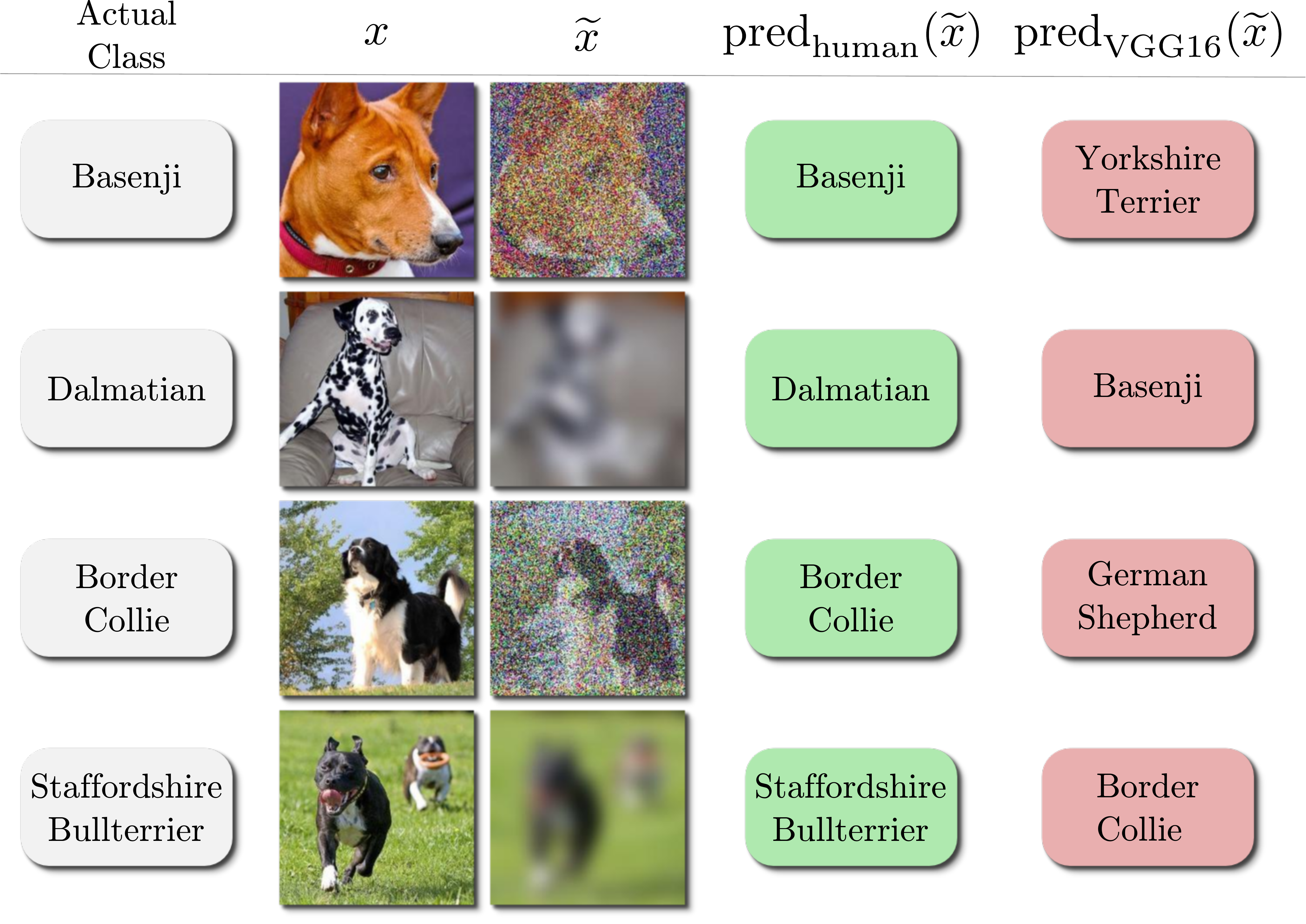} \\
  \caption{\textbf{Human vs DNN predictions on distorted data.} We show several example stimuli from our tests and the prediction given by $>90\%$ of human subjects compared with the prediction from a fine-tuned VGG16 network.}
  \label{fig:tease}
\end{figure}

To answer this question we perform classification experiments with 15 human subjects. We ask the subjects to classify images that are distorted with varied levels of additive Gaussian noise and Gaussian blur. We find that human subjects are able to more accurately classify images under blur and noise distortions compared with DNNs (Figure~\ref{fig:tease}). Furthermore, we find that at high distortion levels the correlation in the errors between deep networks and human subjects is relatively low. This could indicate that the internal models of the DNNs are quite different from the human visual system. These results could be used to guide future research into more robust learning systems. It may be useful to take motivation from the human visual system to achieve good performance on distorted images.

\subsection{Related Works}

Comparing the performance of machine learning systems with human subjects has attracted interest because it may give insights on how machine learning systems can be improved. Borji and Itti \cite{borji-human} compare human classification accuracy and the accuracy of several machine learning algorithms on several datasets. The study considers images as well as line drawings and jumbled images. The study, however, does not test distorted images, and does not include deep learning algorithms.

Fleuret \etal \cite{fleuret} test human and computer performance on a range of synthetic classification tasks. The tasks are binary classification experiments on synthetic images where the difference between the two classes is determined by the spatial arrangement of the constituent parts. Even though these tasks are simple, machine performance is still well below human performance. Stabinger \etal \cite{25years} test modern deep networks on these same classification tasks, but find that the networks can still not achieve human accuracy. While these tests are interesting, the experiments do not represent real data that is likely to occur in most applications.

Parikh \cite{parikh} tested human and computer performance on jumbled images. Jumbled images are constructed by splitting the image into patches and randomly permuting the patches. Global information is lost with jumbling and only local information can be used for classification. The jumbled image can be thought as a distortion of the original image, but it is not a distortion that occurs naturally due to data acquisition or transmission.

The aforementioned studies reach similar conclusions that machine learning algorithms still lack the classification capability of human subjects. However, in the past several years, developments in deep learning have closed this gap and even surpassed human performance. More recent experiments \cite{imagenet} show that human classification accuracy on the large scale ImageNet dataset is less than that of state-of-the-art DNNs (e.g. \cite{resnet}).

Given that DNN performance now can match or exceed human performance, we can begin to examine more difficult problems. Humans have the capacity to recognize severely distorted images. For example, Torralba \etal \cite{tinyimages} show that human subjects can accurately recognize low resolution images. Similarly, Bachmann \cite{faceblur} show that human subjects can recognize low resolution faces. Chen \etal \cite{facenoise} show the effect of noise on a face recognition task. Performance is impaired, but subjects are still able to perform the task with reasonable accuracy. Given that humans can recognize to some degree under quality distortions, it is interesting if DNNs can show the same ability.

However, it has recently been shown that deep networks do not achieve good performance on distorted images \cite{dodge}. Noise and blur were shown to affect the DNNs the most compared with other distortions.

It is not surprising that deep networks trained on clean images perform poorly on distorted images. DNNs are optimized using the statistics of a collection of images, and when the statistics in the testing stage are very different, the model cannot generalize well. Most efforts to add robustness involve fine-tuning DNNs on distorted images. Vasiljevic \etal \cite{blurnetworks} show that fine-tuning yields improvements for blurred images. Similarly, Zhou \etal \cite{distort-icassp} show that this approach also works for images distorted with noise. However, Dodge and Karam \cite{my-iccv-paper} show that models fine-tuned on one type of distortion do not easily generalize to other types of distortions. They show how a gating network can be used to determine the distortion type and level of an input, and pass the data to an appropriately trained network. Finally, the dirty pixel approach \cite{dirty-pixel} shows that fine-tuning with an additional pre-processing module can yield a DNN more robust to blur and noise.

Is fine-tuning the best we can do? We can expect some dropoff in performance when distortions are added because information is lost. But, is the dropoff mostly due to the loss of information? Or are there some inherent properties of existing DNNs that make it difficult to recognize signal from noise?

It may be difficult to analytically answer these questions. Nevertheless, we can experimentally compare DNNs with another visual recognition system, namely the human visual system, and see whether a similar dropoff in performance can be observed. To this aim, we perform experiments to gauge the performance of human subjects on a classification task under quality distortions. By comparing these results with the results from DNNs, we can perhaps gain insight into how to build deep networks that are more robust to distortions.

\section{Methods}

\begin{figure*}[!tb]
  \centering
  \ig{0.9\textwidth}{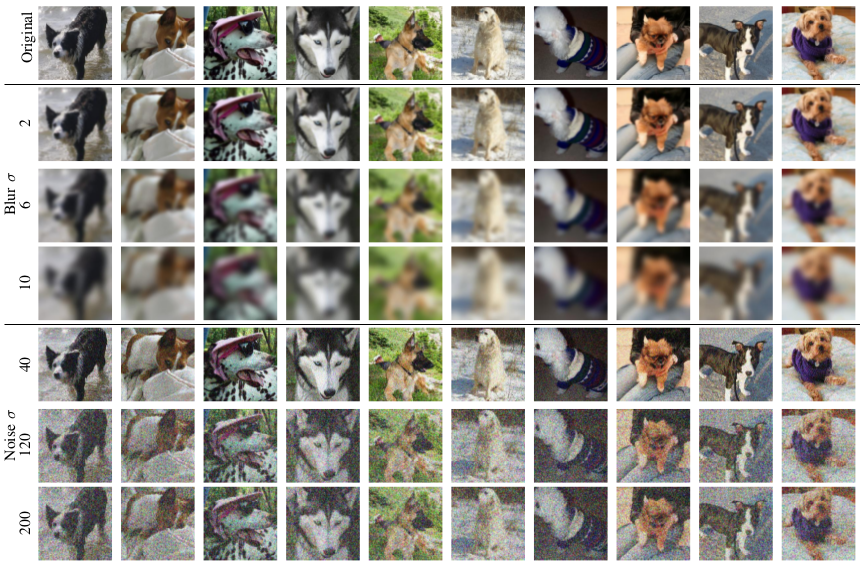}
  \caption{\textbf{Example stimuli.} We show samples from each of the 10 classes of the dataset as well as the corresponding distorted images for three levels of noise and three levels of blur.}
  \label{fig:stimuli}
\end{figure*}

\subsection{Dataset}
Ideally, we would test performance on a large benchmark dataset with many classes (e.g., the 1000 class ImageNet \cite{imagenet}). However, the large number of images and classes presents experimental difficulties. The large number of images would require a significant number of human subjects, or make the experiment very time consuming. Long experiments can induce fatigue in the subjects, which could affect classification performance.

Instead we choose a subset of 10 classes from the ImageNet dataset. We choose the following classes of dogs: border collie, Eskimo dog, German shepherd, pekinese, staffordshire bullterrier, yorkshire terrier, basenji, dalmatian, golden retriever, and miniature poodle. We specifically choose classes of dogs instead of random classes, because we want the classes to be related such that they are difficult to recognize. Note that because we take this data from the ImageNet dataset, there may be some mislabeled data or images that contain more than one dog. We do not filter the data, and assume that the majority of images are correctly labeled with the dominant object in the image.

We consider two types of distortions in this study: Gaussian blur and additive Gaussian noise. It was previously shown in \cite{dodge} that these types of distortions yield the most degradation on DNN classification performance. For Gaussian blur we test 5 different standard deviation values from 2 to 10. For additive noise we use 5 different standard deviation values from 40 to 200. The maximum of the noise range is much higher than previous experiments, because we found in preliminary testing that human subjects still achieve high classification accuracy at a standard deviation of 100. Figure~\ref{fig:stimuli} shows examples of the stimuli from the 10 classes for three levels of the two distortions.

For the experiments, we took images from the validation set of the ImageNet dataset. We need to limit the number of images to be relatively small such that the human study can be completed in a reasonable time. From the possible 50 images per class, we take 25 images per class for training, 5 images per class for validation, and the remaining 20 images per class for testing. With 10 classes, this makes 250 training images, 50 validation images, and 200 testing images. For the 20 test images per class, we apply five levels of blur to 10 images per class, and apply five levels of noise to the remaining 10 images per class. Including all of the distorted versions for all of the classes and the corresponding clean images, there are a total of 1200 testing images available to be used for the experiments.

\subsection{Human Experiments}

\begin{figure}[!tb]
  \centering
  \includegraphics[width=0.4\textwidth]{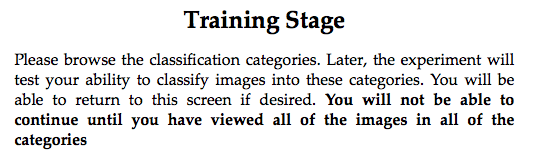}
  \includegraphics[width=0.5\textwidth]{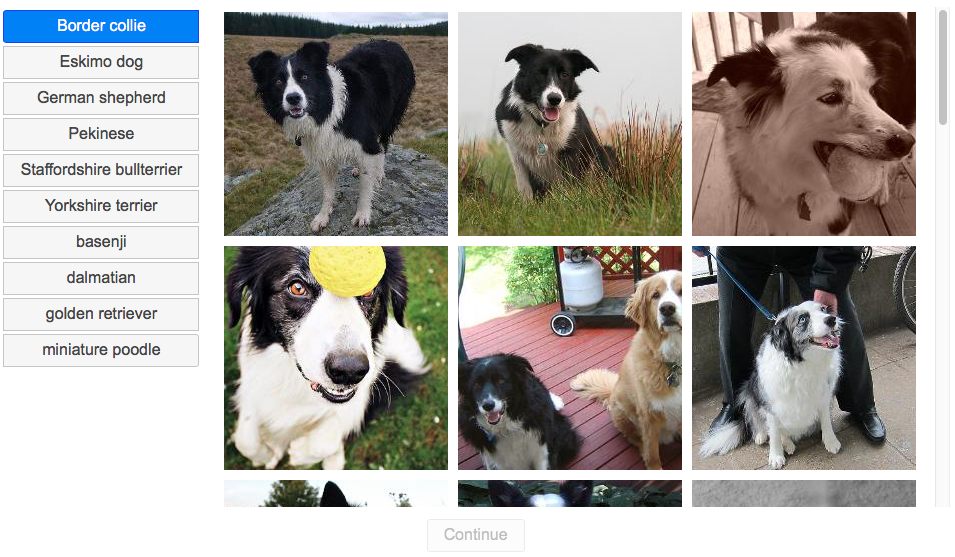}
  \caption{\textbf{Human experiment training procedure.} The training procedure allows the subject to freely view the images in the training set. The subject must view all of the training images in all of the classes before allowed to continue. This is enforced by disabling the continue button until all the images have been viewed.}
  \label{fig:training}
\end{figure}

The experiment begins with a training procedure (Figure~\ref{fig:training}). The subject is allowed to freely view the training images from the 10 classes. The subject must view every training image from each class before the subject can continue to the next part of the experiment.

Next, we perform a validation procedure where the subject classifies clean images. The validation images are shown in a random order. We only allow the subject to choose from the 10 class labels. In other words, we do not offer a ``don't know'' option. We want to force the subject to make their best guess, even if the subject is not confident about the classification. This is analogous to the DNN outputs which will always predict one particular class, even if that class prediction has low confidence. The results of the validation stage establish a baseline performance for the human subjects. Additionally, we can remove the data from outlier subjects that perform poorly on the validation set relative to all other subjects, indicating that the subject is not diligently undertaking the experiment.

Finally in the testing procedure, we gauge the subjects' ability to classify distorted images. In the design of this stage, we take care to avoid a memory effect. If the subject views an image, and then later views the same image again with distortion, the subject may be able to ``fill in'' the missing information with the information from memory. At the same time, it might be useful to test on a single image for what level of distortion the image becomes recognizable.
Thus in our experiment, when a new image is shown, it is shown first with the highest level of distortion. If the subject is able to correctly classify this image, we make the assumption that the subject would also correctly classify the same image under reduced distortion levels. This assumption reduces the total number of images to be tested. Showing all possible 1200 testing images is not feasible for a single subject. If the subject is not able to classify the image at the highest level of distortion, then the level of distortion is reduced and the image is shown again later. This process continues until the subject is shown the clean version of the image. Because we show images from high distortion levels to low distortion levels, we mitigate any potential memory effect. The testing images are shown in a random order to remove any potential effects from the sequential arrangement of images.

\begin{figure}[!tb]
  \centering
  \includegraphics[width=0.4\textwidth]{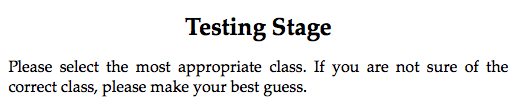}
  \includegraphics[width=0.5\textwidth]{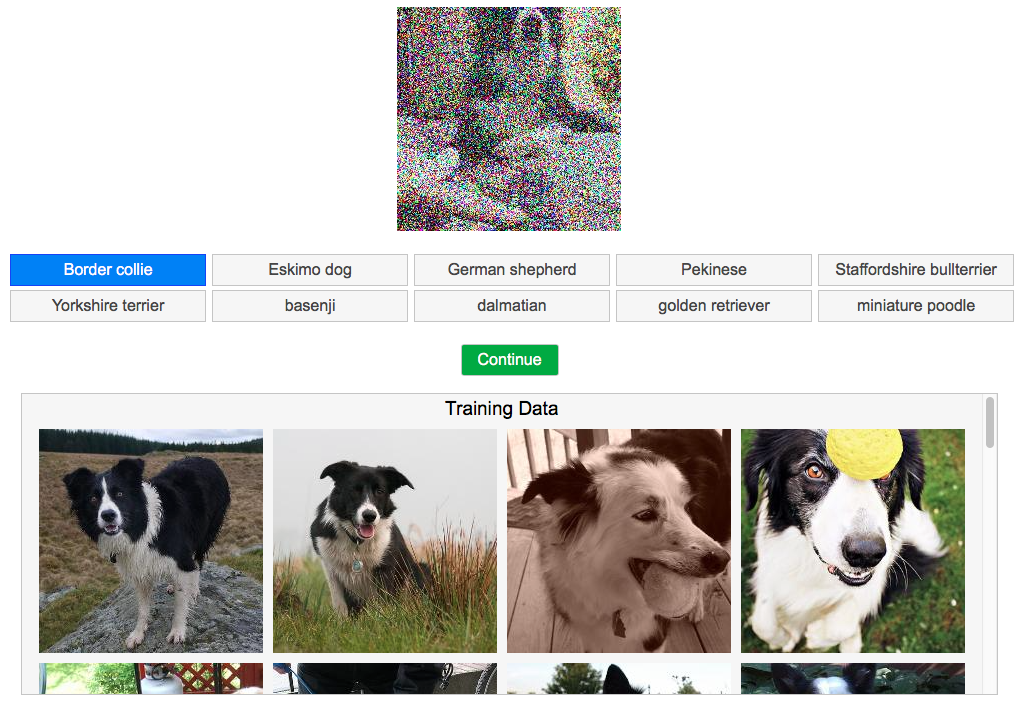}
  \caption{\textbf{Human experiment testing procedure.} The subject is shown a distorted image and asked which class he/she thinks the image best belongs to. After clicking on a class the subject can view the associated training examples. The clicked class is selected as the best choice when the subject clicks on the ``Continue'' button.}
  \label{fig:testing}
\end{figure}

The validation and testing interface is shown in Figure~\ref{fig:testing}. At the top of the screen we show the image under test and instructions to choose the most appropriate class. As the subject hovers the mouse over a particular class or clicks on a class, the associated training images are shown in the bottom panel. This is to ensure that we are testing the ability of the subject to perform recognition, rather than relying on the memory of the subject from the training images. The subject selects a class to be the best one by clicking on the class and then clicking on the ``Continue'' button to go to the next image. The experiment concludes when all of the images have been shown.

We recruited 15 subjects using the Amazon Mechanical Turk crowd-sourcing platform\footnote{\url{http://mturk.com}}. We require that the window size used to view the experiment is large enough to view the entire testing interface without scrolling. This is enforced automatically by our web-based graphical user interface. If the window size is too small, the subject will not be able to continue with the experiment. This restriction also serves to prevent subjects from using mobile devices to take the experiment. 

\subsection{Deep Neural Networks}
A deep neural network (DNN) is a deep layered structure of computational neurons. The neurons are controlled by parameters that can be learned to optimize an objective function using gradient descent algorithms. We consider deep networks that are trained to solve classification problems. A more detailed introduction to deep neural networks can be found in \cite{bengiobook}.

\begin{figure*}[!htb]
  \centering
  \includegraphics[width=0.95\textwidth]{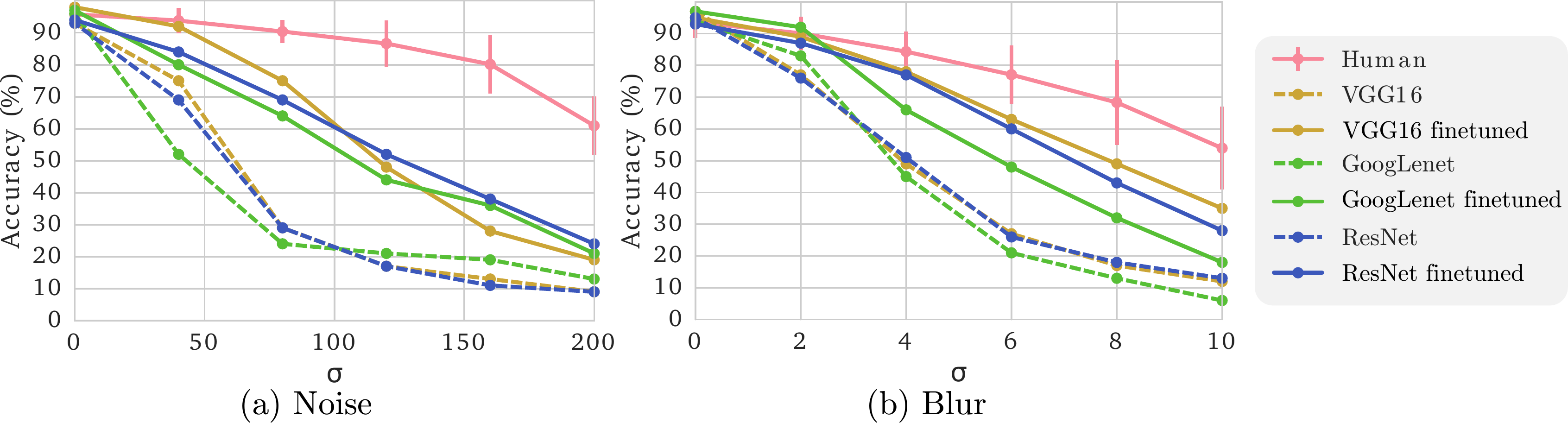}
  \caption{\textbf{Average accuracy on distorted images.} Human subjects outperform deep neural networks on blurred and noisy images, even if the networks are fine-tuned on the respective distortions.}
  \label{fig:accuracy}
\end{figure*}

We consider 3 deep neural network architectures: VGG16 \cite{vgg}, Google Inception version 3 \cite{google}, and the 50 layer ResNet model \cite{resnet}. Each of these networks was pre-trained on the full 1000 classes of the ImageNet dataset. We replace the last layer of these networks with a smaller layer of 10 units to predict the 10 classes of dogs. We train the parameters of this new layer on the clean training images from our dataset subset. We stop training once the loss on the validation set does not decrease.

Following the results of \cite{blurnetworks} and \cite{distort-icassp}, we additionally fine-tune the networks on distorted data. We separately fine-tune networks on noisy or blurred data. During fine-tuning, half of the images from each batch are distorted with a distortion level chosen randomly from a uniform distribution from 0 to the maximum distortion level. The remaining half of the batch is left undistorted. This is to ensure that the DNN can maintain good performance on clean images.

We test the DNNs using the same procedure as for the human studies. We start with the highest level of distortion for each image. If the network correctly predicts the distortion, then we assume that the network can correctly predict the same image at smaller distortion levels. If the network misclassifies an image, then we test the image again at a reduced level. We test the networks in this way such that there is no advantage for the human subjects.


\section{Results}
The subjects achieved an average classification accuracy of $92.9 \pm 5.2 \%$ on the clean validation images. We removed the data from one subject that achieved only a 8\% validation accuracy. The median time spent on each validation image was 6.01 seconds, which shows that the test subjects are diligently performing the experiment. Additionally, the median time spent on the training portion was 10.86 minutes.

Figure \ref{fig:accuracy} shows a comparison of the classification accuracy for the human subjects and the deep network models. The error bars show plus and minus one standard deviation. We see that, even with our sample size of 15 subjects, the human subject results are relatively consistent. Compared with the original deep network models, fine-tuning shows increased accuracy, but accuracy is still much lower than human performance. Human performance drops off quicker for blurred images compared with noisy images, but deep neural networks show the opposite trend.

\begin{figure*}[!tb]
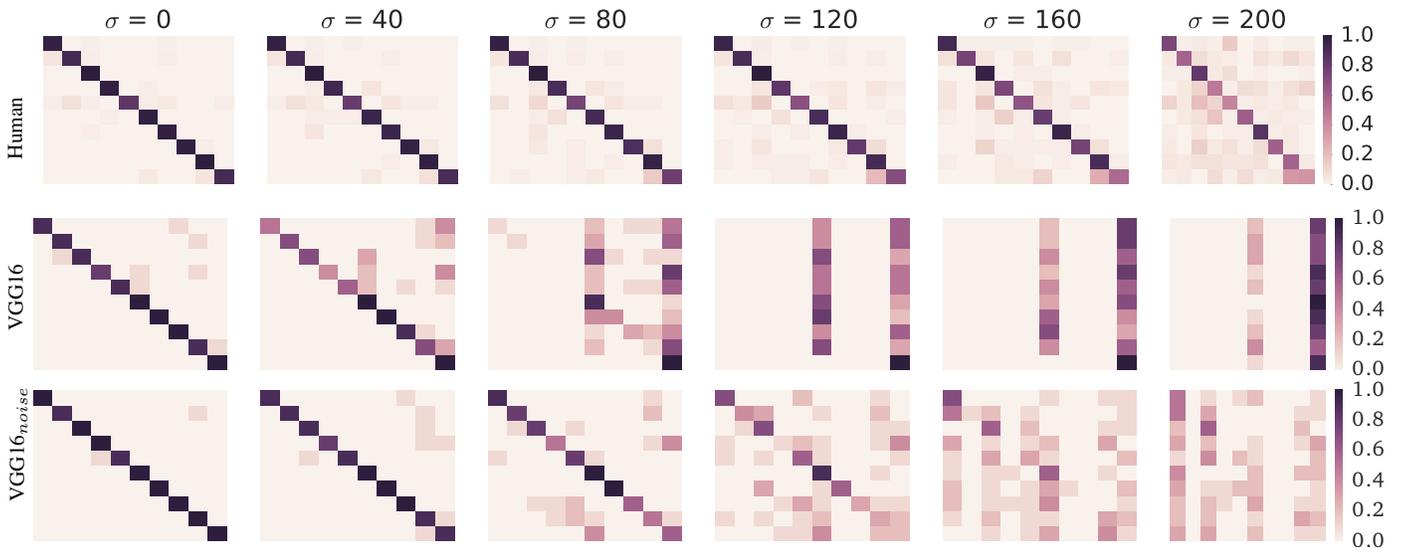

  \centering
  \footnotesize
\setlength{\tabcolsep}{1pt}
\renewcommand{\arraystretch}{0.8}
  \begin{tabular}{cc}
  \rot{\hspace{1.8em} Human} &
  \ig{0.99\textwidth}{confus/humannoise_confus.pdf} \\
  \rot{\hspace{1.8em} VGG16} &
  \ig{0.99\textwidth}{confus/VGG16_cleannoise_confus2.pdf} \\
  \rot{\hspace{1.8em} $\text{VGG16}_{noise}$} &
  \ig{0.99\textwidth}{confus/VGG16_noisenoise_confus2.pdf} \\
    \end{tabular}
  \caption{\textbf{Confusion matrices on noisy images.}}
  \label{fig:confus_noise}
\end{figure*}

Next we consider the question: do human subjects and DNNs make similar errors? Figures \ref{fig:confus_noise} and \ref{fig:confus_blur} show the confusion matrices for the experiments at different distortion levels. A confusion matrix shows the fraction of images from a particular class that are predicted as other classes. Interestingly at high distortion levels, the original networks tend to classify images into mostly 1 or 2 classes. The distortions for these models can be seen as a type of universal adversarial perturbation \cite{universal-adversarial} that forces the model to predict a particular class for any input. The corresponding fine-tuned networks do not exhibit this problem to the same degree.

\begin{figure*}[!tb]
  \centering
  \includegraphics[width=0.99\textwidth]{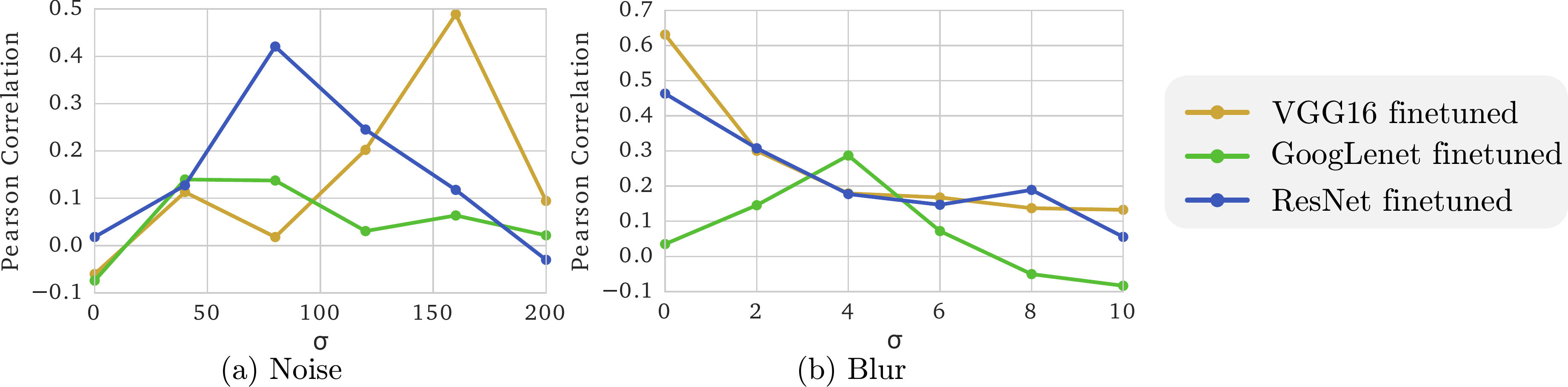}
  \caption{\textbf{Pearson correlation coefficient between the errors of deep learning models and human subjects.} The errors between DNN models and human subjects do not exhibit significant correlations.}
  \label{fig:correlation}
\vspace{20pt}
\end{figure*}

With the confusion matrices we can compute the Pearson correlation coefficient between the errors as in \cite{borji-human}. We take the confusion matrices and remove the diagonal, which leaves only the misclassifications. Figure~\ref{fig:correlation} shows the correlation between human and DNN misclassifications for the different distortion levels. In general, human misclassifications and DNN misclassifications do not show a significant degree of correlation. We note that, for low distortion levels, the correlation computation may not be reliable because of the low number of samples misclassified by the human subjects.

\begin{figure}[!tb]
  \centering
  \begin{subfigure}[b]{0.45\textwidth}
      \includegraphics[width=0.23\textwidth]{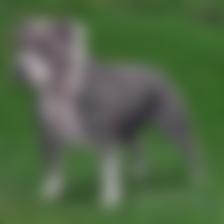}
      \includegraphics[width=0.23\textwidth]{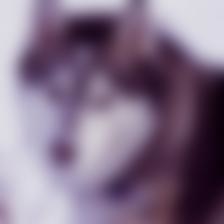}
      \includegraphics[width=0.23\textwidth]{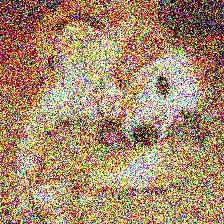}
      \includegraphics[width=0.23\textwidth]{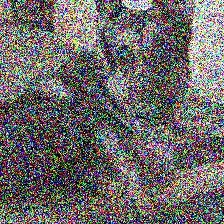}
      \caption{Correct by Humans; Incorrect by VGG16}
  \end{subfigure}

  \begin{subfigure}[b]{0.45\textwidth}
      \includegraphics[width=0.23\textwidth]{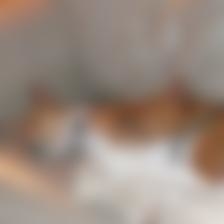}
      \includegraphics[width=0.23\textwidth]{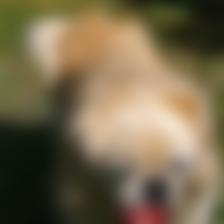}
      \includegraphics[width=0.23\textwidth]{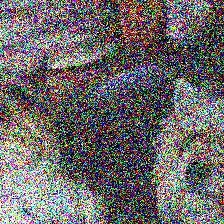}
      \includegraphics[width=0.23\textwidth]{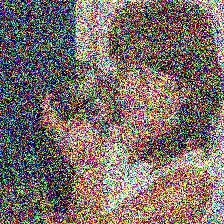}
      \caption{Incorrect by Humans; Correct by VGG16}
  \end{subfigure}

  \begin{subfigure}[b]{0.45\textwidth}
      \includegraphics[width=0.23\textwidth]{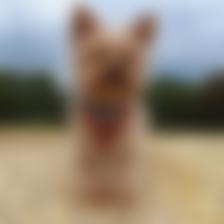}
      \includegraphics[width=0.23\textwidth]{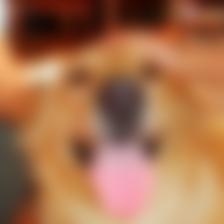}
      \includegraphics[width=0.23\textwidth]{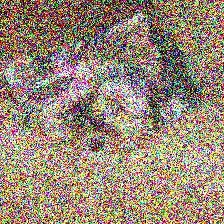}
      \includegraphics[width=0.23\textwidth]{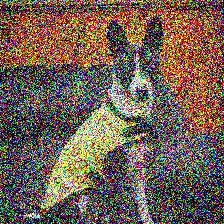}
      \caption{Correct by Humans; Correct by VGG16}
  \end{subfigure}

  \begin{subfigure}[b]{0.45\textwidth}
      \includegraphics[width=0.23\textwidth]{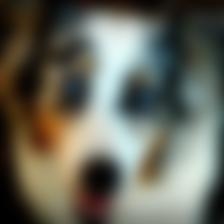}
      \includegraphics[width=0.23\textwidth]{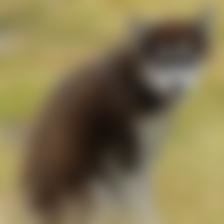}
      \includegraphics[width=0.23\textwidth]{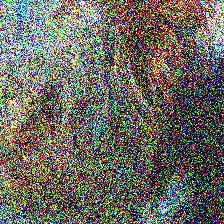}
      \includegraphics[width=0.23\textwidth]{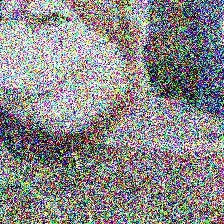}
      \caption{Incorrect by Humans; Incorrect by VGG16}
  \end{subfigure}
  
  \caption{\textbf{Difficult and easy images.} (a) images that at least 90\% of the human subjects classify correctly, but the VGG16 DNN classifies incorrectly. (b) images that the VGG16 DNN classifies correctly, but at least 90\% of the human subjects misclassified. (c) images that are classified correctly by both at least 90\% of the human subjects and the VGG16 DNN. (d) images that are consistently misclassified by both at least 90\% of human subjects and the VGG16 DNN.}
  \label{fig:diff}
\end{figure}

Finally, in Figure~\ref{fig:diff} we show examples of difficult and easy images. Some images are correctly classified by both humans and deep networks, even under heavy distortion. Other images are difficult for both. Some images are easy to recognize by humans, but not DNNs, and other images are the opposite. Although it is difficult to draw conclusions based on these images, it is interesting to note the relative difficulty of different image stimuli.

\begin{figure*}[!tb]
  \centering
  \footnotesize
\setlength{\tabcolsep}{1pt}
\renewcommand{\arraystretch}{0.8}
  \begin{tabular}{cc}
  \rot{\hspace{1.8em} Human} &
  \ig{0.99\textwidth}{confus/humanblur_confus.pdf} \\
  \rot{\hspace{1.8em} VGG16} &
  \ig{0.99\textwidth}{confus/VGG16_cleanblur_confus2.pdf} \\
  \rot{\hspace{1.8em} $\text{VGG16}_{blur}$} &
  \ig{0.99\textwidth}{confus/VGG16_blurblur_confus2.pdf} \\
    \end{tabular}
  \caption{\textbf{Confusion matrices on blurred images.}}
  \label{fig:confus_blur}
\end{figure*}

\section{Discussion and Conclusion}

We show that the performance of DNNs and human subjects is roughly the same on clean images, but for distorted images human subjects achieve much higher accuracy than DNNs. Fine-tuning improves network performance under distorted data, however the performance is still well below that of human subjects. How to close this performance gap is very much an open research question.

Based on our results, it seems that the human visual system has a more robust representation of visual stimuli than state-of-the-art neural networks. Recognition under noise requires that the visual system can still understand the image using global information. Filters in convolutional neural networks may be looking for textures, which becomes difficult under noise.

If the human visual system has a more robust representation, why do DNNs outperform humans on many tasks? It may be because of the broader associative memory of networks. It is very difficult for a human subject to choose a single class given a large number of possible classes. For example, human classification tests on the ImageNet dataset require on average 1 minute per image \cite{imagenet}. Human performance in these types of large scale experiments is also likely reduced due to fatigue.


It is difficult to isolate human and DNN classification accuracy because they both rely on prior experience. For the human subjects, prior experience consists of previous visual exposure to dogs. The human subjects also have prior experience with noisy or blurred stimuli. For the deep networks, prior experience consists of the pre-trained weights learned from the ImageNet dataset and the blurred and noisy images used for fine-tuning.




A potential limitation of our study is the limited number of images and the limited number of classes. We do not expect a larger experiment to significantly change the conclusions we reach in this work. However, a larger experiment could yield more reliable estimations of correlations between the errors of human subjects and DNNs.

\vspace{-8pt}
\section*{Acknowledgment}
The authors would like to thank NVIDIA Corporation for the donation of a TITAN X GPU used in these experiments.

\bibliographystyle{IEEEtran}
\bibliography{refs}
\end{document}